\documentclass[journal]{IEEEtran}
\usepackage{amsfonts}
\usepackage{algorithmic}
\usepackage{algorithm}
\usepackage{array}
\usepackage[caption=false,font=normalsize,labelfont=sf,textfont=sf]{subfig}
\usepackage{textcomp}
\usepackage{stfloats}
\usepackage{url}
\usepackage{verbatim}
\usepackage{graphicx}
\usepackage{cite}
\usepackage{booktabs}
\usepackage{multirow}
\usepackage{amsmath}
\usepackage[normalem]{ulem}
\useunder{\uline}{\ul}{}
\usepackage{hyperref}
\usepackage{booktabs}
\usepackage{multirow}
\usepackage[table]{xcolor}
\usepackage[table]{xcolor}
\allowdisplaybreaks
\usepackage[table]{xcolor}
\usepackage{amssymb}
\usepackage{amsmath}
\usepackage{booktabs}
\usepackage{multirow}
\usepackage{paralist,bm}
\usepackage {amssymb}

\UseRawInputEncoding

\begin{document}

\title{Bridging the Sampling Distribution Shift in Radio Map Estimation: A Trajectory-Aware Paradigm}
% \title{Impact of Sampling Distribution Shift on Learning-Based radio map estimation: A Trajectory-Aware Perspective}
% \title{Rethinking Sampling Strategies for Learning-Based Radio Map Estimation: From Random to Trajectory-Aware}

\author{
Feng Qiu, Zheng Fang, Shuhang Zhang,  \IEEEmembership{Member, IEEE}, Kangjun Liu, Longkun Zou, \\Jing Liu, \IEEEmembership{Senior Member, IEEE}, Ke Chen, \IEEEmembership{Member, IEEE}%
\thanks{This work is supported in part by the Major Key Project of Pengcheng Laboratory under Grant No. PCL2025A02 and PCL2025A14. \textit{(Corresponding authors: Kangjun Liu and Ke Chen.)}
}
\thanks{F. Qiu is with the School of Artificial Intelligence,  Xidian University, Xi’an, China, and also with the Pengcheng Laboratory, China (email: f\_day999@163.com).

Z. Fang is with the Pengcheng Laboratory, China, and also with the Department of Computer Science and Engineering, Southern University of Science and Technology, Shenzhen, China (email: fangzh01@pcl.ac.cn).

S. Zhang is with the Department of Electronics, Peking University, Beijing, China (email: zhangshuhang@pku.edu.cn). 

K. Liu, L. Zou and K. Chen are with the Pengcheng Laboratory, China (email: liukj@pcl.ac.cn; zoulk@pcl.ac.cn; chenk02@pcl.ac.cn); 

J. Liu is with the School of Artificial Intelligence, Xidian University, Xi’an, China, and also with the Guangzhou Institute of Technology, Xidian University, Guangzhou, China (email: neouma@mail.xidian.edu.cn).}

% \thanks{Corresponding authors: Kangjun Liu (email: liukj@pcl.ac.cn) and Ke Chen (email: chenk02@pcl.ac.cn).}%
% \thanks{Kangjun Liu and Ke Chen are corresponding authors (e-mail: liukj@pcl.ac.cn; chenk02@pcl.ac.cn).}%
}

\markboth{Journal of \LaTeX\ Class Files,~Vol.~14, No.~8, August~2021}%
{Shell \MakeLowercase{\textit{et al.}}: A Sample Article Using IEEEtran.cls for IEEE Journals}


\maketitle
\begin{abstract}

Learning-based radio map estimation (RME) plays a critical role in UAV-assisted wireless sensing, enabling tasks such as coverage prediction and network optimization. 
Most current methods assume an independently and identically distributed (i.i.d.) training and testing setting based on random sampling. 
However, practical UAV measurements are collected sequentially along feasible trajectories, resulting in highly structured and spatially correlated patterns. 
This mismatch introduces a sampling distribution shift that increases the intrinsic difficulty of spatial field recovery and compromises the generalization of models trained under i.i.d. assumptions. 
To mitigate this issue, we propose a trajectory-aware training paradigm based on Stochastic-Triggered Trajectory-Based Sampling (ST-TBS), which preserves trajectory continuity while introducing sampling variability. 
Moreover, from a statistical perspective, we show that trajectory-based sampling reduces spatial diversity and increases information redundancy compared to random sampling.
Extensive experiments on the RadioMapSeer and SpectrumNet datasets demonstrate that models trained with random sampling suffer significant performance degradation under trajectory-based observations, with RMSE increasing from 0.0391 to 0.2632 on SpectrumNet. 
Conversely, our proposed ST-TBS method effectively reduces the RMSE to 0.0571. 
These results highlight the necessity of aligning training and deployment sampling distributions for reliable RME.

\end{abstract}


\section{Introduction}
\begin{figure}[t!]
\centering
\includegraphics[width=3.4in]{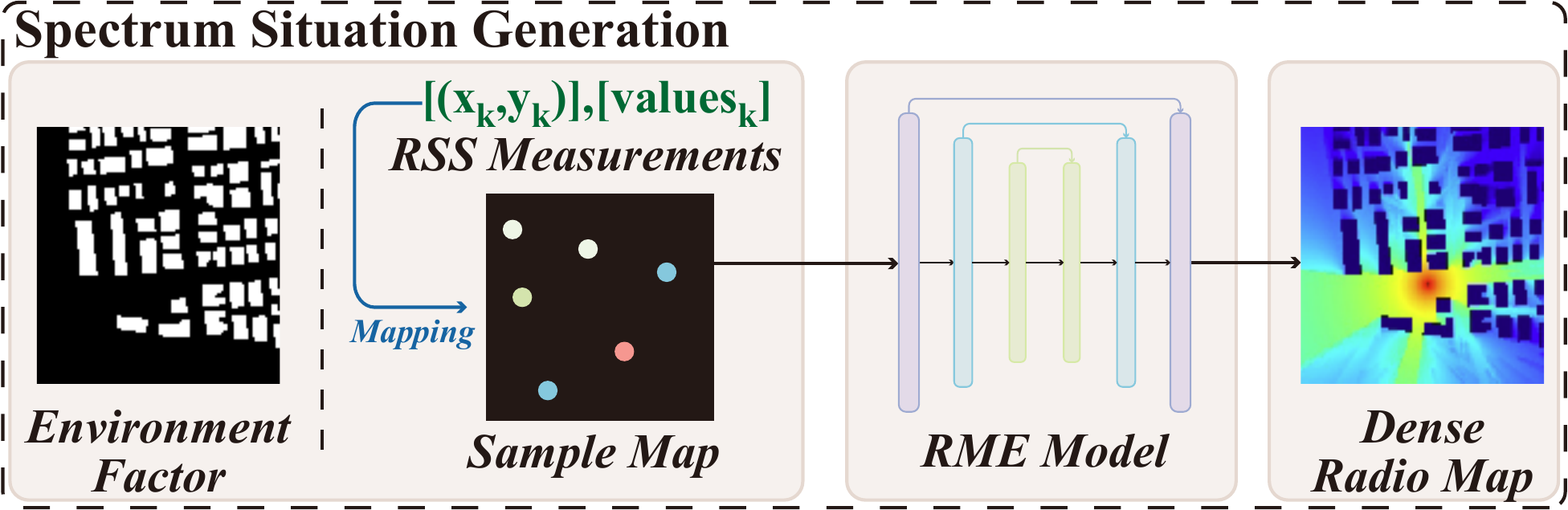}
\caption{The general framework of the RME model, which integrates environmental factors and sparse measurements to reconstruct the radio map.}
\label{fig_1}
\end{figure}

\IEEEPARstart{T}{he} radio map characterizes the spatial distribution of electromagnetic signal strength across a geographic area, serving as a fundamental prerequisite for various wireless communication applications~\cite{3dconstrained, radiomap1}. 
As wireless environments grow increasingly complex, maintaining high-fidelity spectrum situational awareness has become a cornerstone of next-generation network management, facilitating critical tasks such as interference coordination and resource optimization~\cite{spectrum1, spectrum2}. 
However, acquiring a dense, global radio map via direct measurement is often economically and practically prohibitive. Due to hardware costs and physical accessibility constraints, observations are typically limited to a sparse set of locations, yielding discrete Received Signal Strength~(RSS) measurements. 
To bridge the gap between these sparse observations and the necessity for a continuous, high-resolution representation, Radio Map Estimation~(RME) has emerged as an essential solution. 
As illustrated in Fig.~\ref{fig_1}, the RME task aims to reconstruct a high-fidelity, dense radio map by leveraging environmental factors to interpolate and recover the global spectrum distribution from sparse RSS observations.

Recent years have witnessed a rapid evolution in RME techniques, primarily driven by architectural innovations aimed at enhancing model accuracy and efficiency. 
Pioneering works, such as \textit{RadioUNet}~\cite{RadioUNet}, introduced deep learning to the RME task by leveraging convolutional neural networks to learn complex signal-propagation patterns. 
Subsequently, advanced models like \textit{RadioDUN}~\cite{radiodun} and \textit{PMNet}~\cite{PMNet} incorporated physical constraints and radio propagation models into network designs to improve methodological rigor. 
More recently, state-of-the-art architectures such as \textit{RadioFormer}~\cite{radioformer} have further pushed the boundaries by leveraging dual-stream transformers to fuse multi-grained features between building geometries and signal correlations. 
% Despite these remarkable architectural advancements, existing models operate almost exclusively under the idealized assumption of random sampling. 
% It is striking that the sampling strategy---a fundamental component determining how sparse measurements are spatially distributed---has been treated largely as a pre-defined constant, leaving its critical impact on the learning-based reconstruction process virtually unexplored.
Despite these architectural advances, current models are largely built upon the idealized assumption of random sampling. The sampling strategy, which critically dictates the spatial distribution of sparse observations, is frequently treated as a fixed, benign component, leaving its impact on learning-based reconstruction insufficiently explored.

\begin{figure*}[t!]
\centering
\includegraphics[width=\textwidth]{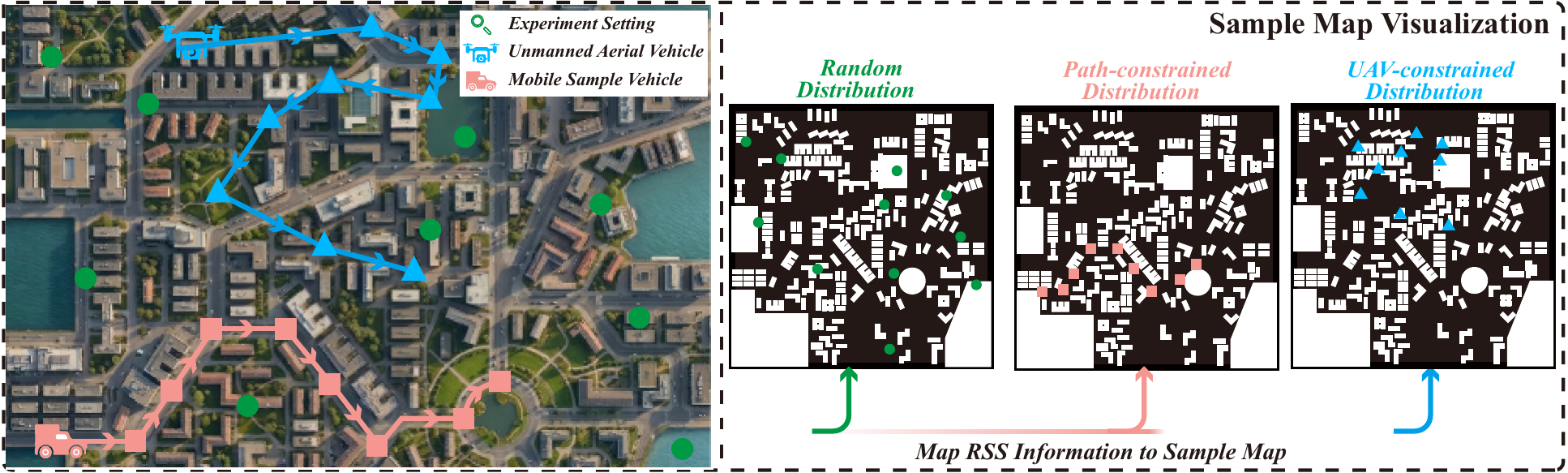}
\caption{Illustration of different sampling strategies in real-world scenarios and their corresponding mapped sample maps, including random distribution (green circles), path-constrained distribution (red squares), and UAV-constrained distribution (blue triangles)~(Best viewed in color).}
\label{fig_2}
\end{figure*}

The necessity to investigate sampling strategies is underscored by previous empirical observations~\cite{radioformer}, which reveal that variations in sampling distributions cause pronounced fluctuations in model performance. 
This exposes a fundamental structural disparity between theoretical assumptions and operational realities. 
As illustrated in Fig.~\ref{fig_2}, existing RME benchmarks predominantly rely on \textit{Random Sampling~(RS)}, assuming that observation points are independently and identically distributed (i.i.d.) across the entire geographic area. 
While this mathematically convenient paradigm provides an unbiased global representation of the electromagnetic field even under extreme sparsity, it is fundamentally detached from practical deployments. 
In real-world urban environments, data acquisition is strictly dictated by physical accessibility and the mobility of sensing hardware. Because locations selected by \textit{RS} often fall into inaccessible areas, such as dense building interiors or restricted zones, practitioners must rely on mobile sampling platforms like vehicles or Unmanned Aerial Vehicles~(UAVs). 
Consequently, these observations are strictly constrained by road networks or kinematic trajectories. This constraint shifts the sampling pattern from a globally uniform state to \textit{Trajectory-Based Sampling~(TBS)}, characterized by strong spatial correlations and localized redundancy. This simulation-to-deployment gap poses a critical challenge, severely degrading the generalization capability and robustness of existing RME models.

% Motivated by this overlooked simulation-to-deployment gap, this work moves beyond the superficial observation of error metrics to systematically investigate the underlying mechanisms of reconstruction failure induced by sampling distribution shifts. 
% We bridge the gap between idealized theory and operational reality through a structured research framework. 
% Initially, we establish an analytical foundation to quantify how trajectory-induced spatial correlations diminish the information gain available to RME models. Subsequently, we conduct a series of experiments across diverse urban datasets to empirically demonstrate the severe performance degradation of randomly trained models under constrained movement.
% Leveraging these insights, we propose a robust mitigation framework that realigns the model's inductive bias with real-world sampling dynamics, ensuring reliable spectrum awareness in physically constrained sensing missions. 

Motivated by this overlooked gap, this work focuses on the fundamental mismatch between idealized i.i.d. sampling assumptions and the trajectory-constrained data acquisition inherent to mobile sensing platforms. 
Moving beyond surface-level performance observations, we systematically investigate the underlying mechanisms of reconstruction degradation caused by this sampling distribution shift. 
To this end, we provide a statistical characterization of how trajectory-induced spatial correlations alter the effective information content available to RME models. Furthermore, we conduct extensive experiments across diverse urban datasets to empirically demonstrate the severe performance drops of randomly-trained models when faced with constrained sampling patterns. 
Building upon these insights, we propose a trajectory-aware training framework that realigns the model’s inductive bias with real-world sampling dynamics, thereby improving robustness in practical deployment scenarios. 

The main contributions of this paper are summarized as follows:
\begin{itemize}
    \item We identify and formalize the \textit{sampling distribution shift} problem in learning-based RME, highlighting the critical mismatch between training-time random sampling and deployment-time trajectory-constrained sampling.
    
    \item We propose Stochastic-Triggered Trajectory-Based Sampling \textit{(ST-TBS)}, a novel training protocol that incorporates random triggering along continuous paths to approximate real-world sampling variability and mitigate the distribution mismatch.
    
    % \item We provide a statistical interpretation of trajectory-based sampling, showing how it introduces strong spatial correlations and alters the effective information structure of observations, which inherently compromises model generalization.
    \item We provide a statistical interpretation of trajectory-based sampling, demonstrating how trajectory-induced spatial correlations reduce spatial diversity and increase information redundancy, which inherently compromises model generalization.
    
    \item We conduct extensive experiments across multiple datasets and state-of-the-art architectures, comprehensively validating the detrimental impact of the sampling shift and demonstrating the robustness and efficacy of the proposed ST-TBS approach.
\end{itemize}

The remainder of this paper is organized as follows: 
Section~\ref{sec:related} reviews related works on learning-based reconstruction, mobility-constrained sensing, and sampling strategies. 
Section~\ref{sec:method} introduces the problem formulation, the concept of trajectory-based sampling, and our proposed TBS-aware training strategy. 
Section~\ref{sec:theory} provides a statistical interpretation of the sampling-induced distribution shift, quantifying information redundancy and generalization risks. 
Section~\ref{sec:exps} presents extensive experimental validation. 
Finally, Section~\ref{sec:conclusion} concludes the paper and discusses future research directions.

\section{Related Works}
\label{sec:related}
Existing research relevant to radio map estimation primarily encompasses three directions: learning-based spatial reconstruction, mobility-constrained UAV deployment, and active sampling strategy design. Despite significant advancements in these areas, existing studies predominantly rely on the idealized assumption of independently sampled measurements, thereby overlooking the highly structured and spatially correlated sampling patterns inherently induced by practical UAV trajectory constraints. In the following subsections, we briefly review these foundational lines of research and highlight their critical limitations concerning this sampling distribution mismatch.

\subsection{Learning-Based Spatial Field Reconstruction}
Spatial field reconstruction has been extensively investigated in wireless sensing and vehicular communication systems. Early approaches relied on model-based techniques such as Kriging interpolation~\cite{Kriging1, kriging2}, Gaussian process regression~\cite{Gaussian1, Gaussian2}, and compressive sensing~\cite{compressive1,compressive2,energy1} to estimate spatial signal strength maps from sparse measurements. These methods provide theoretical guarantees under certain statistical assumptions and are widely used for radio environment mapping and spectrum cartography. 

With the increasing availability of large-scale datasets, learning-based reconstruction methods have become dominant. Convolutional encoder-decoder architectures, such as U-Net variants, have been widely adopted to recover dense spatial fields from sparse observations~\cite{RadioUNet,RadioMamba}. More recently, attention-based and transformer-based models have further improved reconstruction accuracy by capturing long-range spatial dependencies~\cite{attention1,radioformer}. These approaches generally formulate the problem as an image-to-image translation task and consistently achieve superior performance compared to traditional interpolation techniques. 

Despite these advances, both traditional model-based and modern learning-based approaches typically rely on idealized assumptions of independently and randomly distributed sampling locations. In practical UAV-assisted sensing systems, however, sampling locations are determined by physically feasible trajectories rather than independent point selection. This discrepancy between idealized sampling assumptions and motion-constrained data acquisition introduces a fundamental mismatch, which remains largely overlooked in existing reconstruction frameworks.

\subsection{Mobility-Constrained UAV Deployment}
In practical vehicular and UAV-assisted sensing systems, mobility is inherently constrained by physical dynamics, road topologies, energy budgets, and regulatory requirements~\cite{energy1,energy2,energy3,Uncertainty1}. UAV trajectories must satisfy motion continuity, limited acceleration, and feasible navigation corridors. When operating in urban environments, sensing platforms typically follow road networks or predefined flight paths to ensure safety and stability.

Extensive research has focused on trajectory design and energy-aware routing for UAVs, considering factors such as flight endurance, coverage efficiency, and communication quality~\cite{trajectory1,trajectory2,trajectory3}. These studies consistently highlight that UAV sensing is realized through continuous trajectories rather than independent spatial relocations. Even when adaptive planning strategies are deployed, the resulting sampling patterns remain sequential and spatially correlated due to motion feasibility constraints.

While mobility-constrained deployment has been thoroughly examined in the context of coverage optimization and communication performance, its direct implications for learning-based spatial reconstruction have not been systematically explored. In particular, how continuous trajectories reshape the statistical structure of sampling patterns---and consequently affect reconstruction performance---remains unclear.

\subsection{Sampling Strategy Design for Spatial Estimation} 
Sampling strategy design plays a crucial role in spatial field estimation. Active sensing and adaptive sampling techniques aim to sequentially select informative measurement locations to improve spatial estimation efficiency~\cite{active1,active2,active3,active4}. Among these approaches, Bayesian experimental design is widely adopted, where sampling locations are chosen to maximize information gain or reduce posterior uncertainty in probabilistic models such as Gaussian processes~\cite{Uncertainty1,Gaussian2,Uncertainy2}.

In parallel, learning-based sampling strategies have been proposed to jointly optimize sensing policies and reconstruction modules~\cite{learning-based1,learning-based2}. Some works learn sampling masks or point selection policies that are tailored to specific neural reconstruction networks, thereby improving task-oriented performance~\cite{learning-based3}. These approaches demonstrate the potential benefits of integrating sampling design with downstream reconstruction tasks.

However, most existing sampling strategies still assume unrestricted access to spatial locations, selecting sampling points independently from a candidate pool. In contrast, UAV-assisted sensing systems are inherently restricted by continuous trajectories and motion constraints. Such path-based sampling introduces strong spatial correlations between neighboring measurements, leading to information redundancy that is not explicitly modeled in conventional sampling frameworks.

\vspace{0.5em}
\noindent
\textbf{Summary and Motivation:} Prior research has extensively investigated spatial reconstruction methods, UAV mobility constraints, and sampling strategy designs. However, these aspects are largely studied in isolation. To simplify algorithm development, many prior RME studies adopt idealized simulation settings where sampling locations are randomly generated, without explicitly accounting for travel costs or motion feasibility. When reconstruction models trained under these random distributions are deployed in trajectory-constrained scenarios, a severe distribution shift occurs in the spatial structure of the observed measurements. This structural mismatch between theoretical assumptions and practical deployment remains underexplored. This fundamental gap motivates the present study, which aims to systematically analyze continuous trajectory-induced sampling patterns and mitigate their detrimental impact on learning-based spatial reconstruction.

\section{Methodology}
\label{sec:method}

\subsection{Problem Formulation}
We define the radio map as a discretized spatial field distributed over a grid comprising $d$ cells. Let $\mathbf{x} \in \mathbb{R}^d$ denote the vectorized global spectrum map to be estimated. To facilitate theoretical analysis, we model the underlying field using a Gaussian Random Field prior~\cite{GRF}, expressed as $\mathbf{x} \sim \mathcal{N}(\mathbf{0}, \mathbf{K})$. Here, $\mathbf{K} \in \mathbb{R}^{d \times d}$ represents the covariance matrix induced by an isotropic kernel, which characterizes the spatial correlations inherent in electromagnetic environments.

The data acquisition process is modeled as a linear, noisy measurement. Consider a sampled index set $\mathcal{S} = \{s_1, \dots, s_m\} \subseteq \{1, \dots, d\}$ with cardinality $m$. 
We further model the data acquisition process as a probability distribution $\mu$ over sampling index sets $\mathcal{S}$, such that $\mathcal{S} \sim \mu$. In particular, we consider different sampling distributions corresponding to independent random sampling (RS) and trajectory-based sampling (TBS).
We define a selection matrix $\mathbf{P}_{\mathcal{S}} \in \mathbb{R}^{m \times d}$ that extracts the entries from $\mathbf{x}$ corresponding to the indices in $\mathcal{S}$. 
The resulting observation vector $\mathbf{y} \in \mathbb{R}^m$ is expressed as:
\begin{equation}
\label{eq:obs}
\mathbf{y} = \mathbf{P}_{\mathcal{S}} \mathbf{x} + \mathbf{n}, \quad \mathbf{n} \sim \mathcal{N}(\mathbf{0}, \sigma^2 \mathbf{I}_m),
\end{equation}
where $\mathbf{n}$ represents additive white Gaussian noise. 

While Eq.~\ref{eq:obs} characterizes the standard reconstruction problem given a sampling set $\mathcal{S}$, our primary interest lies in the data acquisition process that determines how $\mathcal{S}$ is generated. Specifically, we model the sampling mechanism as a distribution μ over the spatial domain, encompassing both independent random sampling (RS) and trajectory-based sampling (TBS). The goal of this work is to analyze how different sampling distributions μ influence the statistical structure of observations and, consequently, the reconstruction performance of learning-based models.

The RME model, denoted by $f_\theta$, aims to reconstruct the global spatial spectrum field $\widehat{\mathbf{x}} = f_\theta(\mathbf{y}, \mathcal{S}, \mathbf{M})$ from sparse observations and environmental context $\mathbf{M}$. To quantify the reconstruction performance, we define the risk $\mathcal{R}_{\mu}(\theta)$ under a specific sampling distribution $\mu$ as the expected Mean Squared Error (MSE):
\begin{equation}
\label{eq:risk}
\mathcal{R}_{\mu}(\theta) \triangleq \mathbb{E}_{\mathcal{S} \sim \mu} \left[ \mathbb{E}_{\mathbf{x}, \mathbf{n}} \left[ \|\widehat{\mathbf{x}} - \mathbf{x}\|^2 \right] \right],
\end{equation}
where $\mu \in \{p_R, p_C\}$ represents the distributions for \textit{Random Sampling} and \textit{Trajectory-Based Sampling}, respectively. We further denote the domain-optimal parameters as $\theta_R \in \arg\min_\theta \mathcal{R}_{p_R}(\theta)$ and $\theta_C \in \arg\min_\theta \mathcal{R}_{p_C}(\theta)$, representing the minimum achievable risk within the given model class under each sampling distribution when the model is tailored to the specific statistical regularities of \textit{RS} and \textit{TBS}.

\begin{algorithm}[t]
\caption{Generation Procedure ($\mathcal{G}_{TBS}$) of Trajectory-Based Sampling ($\ell$-TBS and ST-TBS)}
\label{alg: tbs}
\begin{algorithmic}[1]
\REQUIRE Environment mask $\mathbf{M}$, Sample number $m$, Persistence probability $P_p$, Trigger probability $P_{trig}$, Mode $\in \{\ell\text{-TBS}, \text{ST-TBS}\}$, Distance Threshold $\ell$.
\ENSURE Sampling set $\mathcal{S}$.
\STATE Initialize $\mathcal{S} = \emptyset$, current location $s_{curr} \in \mathbf{M}_{free}$, previous direction $\mathbf{d}_{prev} = \text{null}$, accumulated distance $d_{acc} = 0$
\WHILE{$|\mathcal{S}| < m$} 
    \STATE \textbf{Direction Selection:} Generate $r \sim \text{Uniform}(0,1)$
    \IF{$r < P_p$ \textbf{and} $\mathbf{d}_{prev} \neq \text{null}$}
        \STATE $\mathbf{d}_t \gets \mathbf{d}_{prev}$ \COMMENT{Keep previous momentum}
    \ELSE
        \STATE $\mathbf{d}_t \sim \text{Uniform}(\{\text{Up, Down, Left, Right}\})$ \COMMENT{Sample new direction}
    \ENDIF
    \STATE \textbf{Movement:} $s_{next} \gets s_{curr} + \mathbf{d}_t$
    \IF{$s_{next} \in \mathbf{M}_{free}$}
        \STATE $s_{curr} \gets s_{next}$, $\mathbf{d}_{prev} \gets \mathbf{d}_t$
        \STATE $d_{acc} \gets d_{acc} + 1$ \COMMENT{Update accumulated travel distance}
        \STATE \textbf{Sampling Trigger:}
        \IF{Mode is ST-TBS}
            \STATE Generate $\xi \sim \text{Uniform}(0,1)$
            \IF{$\xi < P_{trig}$ \textbf{and} $s_{curr} \notin \mathcal{S}$}
                \STATE $\mathcal{S} \gets \mathcal{S} \cup \{s_{curr}\}$ 
            \ENDIF
        \ELSIF{Mode is $\ell$-TBS}
            \IF{$d_{acc} \geq \ell$ \textbf{and} $s_{curr} \notin \mathcal{S}$}
                \STATE $\mathcal{S} \gets \mathcal{S} \cup \{s_{curr}\}$
                \STATE $d_{acc} \gets 0$ \COMMENT{Reset accumulated distance}
            \ENDIF
        \ENDIF
    \ELSE
        \STATE $\mathbf{d}_{prev} \gets \text{null}$ \COMMENT{Reset heading upon collision}
    \ENDIF
\ENDWHILE
\RETURN $\mathcal{S}$
\end{algorithmic}
\end{algorithm}

\subsection{Trajectory-Based Sampling}
To bridge the gap between an idealized research setting and real-world sensing scenarios, we define \textit{Trajectory-Based Sampling} as a constrained stochastic process in which the sampling set $\mathcal{S}$ is generated along a continuous path $\mathcal{P}$. 
Unlike \textit{RS}, where indices are selected independently, \textit{TBS} is governed by the kinematic and environmental constraints of a mobile sensing platform.

The transition between consecutive samples is defined as the following recursive relation:
\begin{equation}
s_{t+1} = \Phi\left(s_t + \mathbf{v}_t \cdot \Delta t, \mathbf{M} \right),
\end{equation}
where $\mathbf{M} \in \{0,1\}^d$ is a binary building mask and $\Phi(\cdot)$ is a projection operator that enforces building avoidance, ensuring $s_{t+1} \in \mathbf{M}_{free}$. 
A fundamental constraint in \textit{TBS} is the kinematic distance threshold $\ell$ between successive observations:
\begin{equation}
\label{eq:distance threshold}
\|s_{t+1} - s_t\| \leq \ell.
\end{equation}
In practical systems, $\ell = v/f$ represents the spatial interval determined by the platform velocity $v$ and the sensing frequency $f$. This kinematic constraint enforces sequential observations to be spatially clustered, introducing localized redundancy and strong short-range spatial correlations. In addition to this local continuity constraint, the trajectory is further restricted by global accessibility conditions, such as road network structures or feasible flight corridors, which limit the reachable regions and traversal patterns of the sensing platform. Together, these constraints reshape the statistical structure of the collected data compared to independent spatial sampling, leading to reduced spatial diversity and increased redundancy.

To operationalize this continuous sensing process, we detail the path generation and measurement acquisition procedure in Algorithm~\ref{alg: tbs}. The underlying trajectory is constructed utilizing a persistent random walk to emulate the kinematic momentum of a mobile platform. \textbf{We do not aim to faithfully reproduce full UAV dynamics; instead, we construct a simplified trajectory-consistent sampling generator that captures the statistical characteristics most relevant to radio map reconstruction, namely spatial continuity and local correlation.} At each sequence step, the platform maintains its previous directional heading with a persistence probability $P_p$, ensuring path continuity while strictly navigating within the collision-free space $\mathbf{M}_{free}$. 

\begin{figure}[t]
\centering
\includegraphics[width=\linewidth]{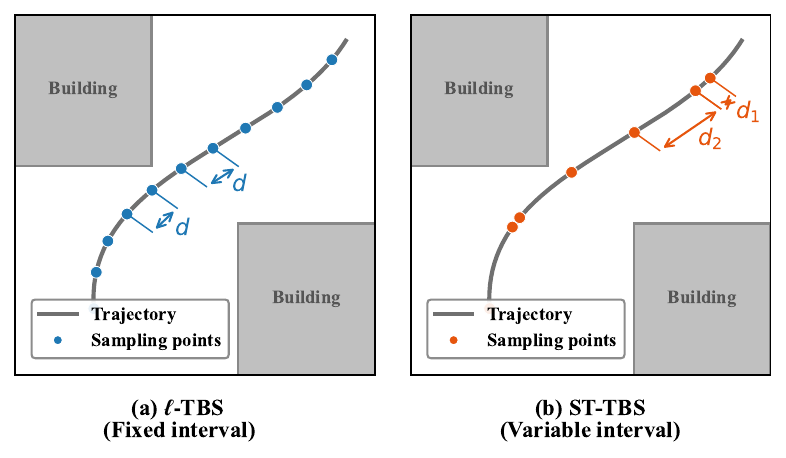}
\caption{Comparison of different triggering mechanisms within the \textit{TBS} framework. (a) \textit{Fixed-Interval Triggering} collects measurements at constant spatial steps, resulting in periodic redundancy. (b) \textit{Stochastic Triggering} introduces random jitter along the trajectory, breaking the periodic pattern to enhance distribution diversity.}
\label{fig:trigger_comparison}
\end{figure}

To extract discrete observations along this generated path, we implement two distinct triggering mechanisms, as illustrated in Fig.~\ref{fig:trigger_comparison}. The \textit{Fixed-Interval Triggering} ($\ell$-\textit{TBS}) records a sample whenever the accumulated travel distance reaches the threshold $\ell$, effectively mimicking standard sensors operating with constant sampling rates and uniform platform velocity. In contrast, the \textit{Stochastic Triggering~(ST-TBS)}, which serves as the core mechanism in Algorithm~\ref{alg: tbs}, evaluates a probabilistic trigger at each valid spatial step. A measurement is recorded independently with a trigger probability $P_{trig}$. This stochastic approach introduces random spatial jitter, effectively breaking the periodic redundancy inherent in fixed-interval paths. Collectively, these mechanisms allow the sampling model to capture the primary statistical properties of trajectory-based sensing---namely, spatial continuity and localized correlation---without necessitating task-specific path planning or complex UAV control dynamics.

\subsection{TBS-Aware Training Strategy}
To mitigate the generalization gap between idealized theoretical assumptions and practical deployment, we propose a \textit{TBS-Aware Training Strategy}. The core principle of this approach is to align the model's inductive bias with the non-i.i.d. nature of trajectory data by integrating the TBS generator directly into the training pipeline. 

Unlike conventional training paradigms that utilize static random masks, our strategy employs a simplified trajectory-consistent sampling generator that captures key statistical properties (continuity and locality).
%a dynamic, on-the-fly sampling approach. 
During each training iteration, for a given ground-truth radio map $\mathbf{x}_i$, a unique sampling set $\mathcal{S}_i$ is dynamically generated via the TBS generator $\mathcal{G}_{TBS}$, as detailed in Algorithm~\ref{alg: tbs}. This procedural integration ensures that the neural network $f_\theta$ is consistently exposed to the localized redundancy and spatial correlations characteristic of real-world UAV sensing. The optimization objective is formulated as:
\begin{equation}
\min_{\theta} \sum_{i=1}^{N} \mathcal{L} \left( f_\theta(\mathbf{P}_{\mathcal{S}_i} \mathbf{x}_i, \mathcal{S}_i, \mathbf{M}_i), \mathbf{x}_i \right), \quad \mathcal{S}_i \sim \mathcal{G}_{TBS}(\mathbf{M}_i),
\end{equation}
where $N$ denotes the batch size, $\mathbf{M}_i$ represents the environmental factors (e.g., building geometries), and $\mathcal{L}$ denotes the reconstruction loss. 

Specifically, we leverage the \textit{ST-TBS} mechanism during the training phase. By introducing stochastic spatial jitter into the continuous trajectory, the model learns to reconstruct the global spatial field from heavily clustered measurements while maintaining robustness to the variable sampling densities inherent in kinematic sensing. Consequently, this strategy compels the network to move beyond simple localized interpolation, fostering a comprehensive structural understanding of the electromagnetic environment under physically constrained observation patterns.

% \section{Theoretical Analysis}
\section{Statistical Interpretation of Trajectory-Based Sampling}
\label{sec:theory}
% In this section, we provide a theoretical framework to explain the performance disparities observed between \textit{RS} and \textit{TBS}. By formulating the kinematic constraints as a Markov process, we analyze how \textit{TBS} inherently induces a spatial short-distance bias. Furthermore, we leverage information-theoretic principles to quantify the resulting redundancy penalty and decompose the generalization risk. This analysis elucidates the fundamental causes behind the severe performance degradation of RS-trained models when evaluated under trajectory-constrained conditions.

In this section, we provide a statistical interpretation of trajectory-based sampling to better understand the impact of sampling mechanisms on reconstruction performance. Rather than deriving strict theoretical guarantees, our goal is to offer insights into how spatial correlation and sampling structure influence the effective information content of observations.

\subsection{Sampling-Induced Distribution Shift and Hypotheses}
The transition from \textit{RS} to \textit{TBS} introduces a fundamental shift in the distribution of the sampling mask space. Let $p_R$ and $p_C$ denote the spatial sampling distributions for \textit{RS} and \textit{TBS} (kinematically constrained), respectively. Let $\mathcal{R}_{\mu}(\theta)$ denote the MSE risk of model $\theta$ evaluated under a specific sampling distribution $\mu$. Based on the empirical gap between i.i.d. observations and path-dependent measurements, we formulate two core hypotheses. 

First, we define \textit{Shift Degradation}, positing that a model $\theta_R$ optimized under random sampling suffers significant performance degradation when evaluated on \textit{TBS} data, mathematically expressed as $\mathcal{R}_{p_C}(\theta_R) > \mathcal{R}_{p_R}(\theta_R)$. Second, we introduce the principle of \textit{Domain Optimality}, which suggests that models exhibit superior generalization when the testing sampling distribution aligns with the training phase. Specifically, we expect that $\mathcal{R}_{p_R}(\theta_R) \leq \mathcal{R}_{p_R}(\theta_C)$ and $\mathcal{R}_{p_C}(\theta_C) \leq \mathcal{R}_{p_C}(\theta_R)$, where $\theta_C$ represents a model trained under trajectory constraints.

\subsection{Kinematic Constraints and Short-Distance Bias}
To bridge the gap between physical movement and statistical redundancy, we model the TBS process as a Markov chain $\{s_t\}_{t=1}^m$ on the spatial grid $\Omega$. The transition kernel is governed by a bounded step size constraint $\| s_{t+1} - s_t \| \leq \ell$. To quantify the degree of spatial clustering among samples, we define a statistic that measures the average pairwise proximity between sampling locations: $T_{\phi}(\mathcal{S}) \triangleq \frac{2}{m(m-1)} \sum_{a<b} \phi(d(s_a, s_b))$,
where ${\phi(\cdot)}$ measures pairwise spatial proximity.

In a finite-horizon setting, the bounded-step nature of \textit{TBS} limits the spatial mixing of samples. For an indicator function $\phi_{r_0}(r) = \mathbf{1}\{r \leq r_0\}$, the expected number of short-distance pairs under \textit{TBS} satisfies:
\begin{equation}
\label{eq:shortpair_bias}
\mathbb{E}_{\mathcal{S} \sim p_C}[T_{\phi_{r_0}}(\mathcal{S})] \geq \mathbb{E}_{\mathcal{S} \sim p_R}[T_{\phi_{r_0}}(\mathcal{S})],
\end{equation}
% “近距离点对更多”
where the equality holds only asymptotically as $\ell \to \infty$. This inequality indicates that, compared to \textit{RS}, \textit{TBS} produces a higher proportion of spatially close sample pairs, reflecting a strong clustering effect.

% This inequality formally establishes that \textit{TBS} inherently induces a clustering bias, in which consecutive observations are confined to a localized neighborhood, in stark contrast to the uniform spatial coverage of \textit{RS}.

% Under simplifying assumptions (e.g., uniform initialization and absence of obstacles), trajectory-based sampling tends to produce a higher proportion of short-distance sample pairs compared to independently drawn random samples, i.e.,
% \begin{equation}
% E_{S∼p_C}[T_{φ_{r_0}}(S)] ≳ E_{S∼p_R}[T_{φ_{r_0}}(S)],
% \end{equation}

% where ${φ\_{r_0}(r) = 1{r ≤ r_0}}$ is an indicator function.

% This relation should be interpreted as an approximate statistical tendency rather than a strict inequality. It reflects the clustering effect induced by trajectory continuity, where consecutive observations are more likely to remain within a localized neighborhood, in contrast to the more uniformly dispersed samples under random sampling.
% This short-distance bias plays a key role in reducing spatial diversity, which will be further analyzed in the following section.

\subsection{Information Redundancy and Log-Det Penalty}
Under the Gaussian prior $\mathbf{x} \sim \mathcal{N}(\mathbf{0}, \mathbf{K})$, the conditional mutual information $I(\mathbf{x}; \mathbf{y} \mid \mathcal{S})$ is characterized by $\frac{1}{2} \log \det (\mathbf{I}_m + \frac{1}{\sigma^2} \mathbf{A}(\mathcal{S}))$, where $\mathbf{A}(\mathcal{S})$ is the covariance submatrix of the sampled locations. To analyze the impact of spatial clustering, we decompose this matrix as $\mathbf{A}(\mathcal{S}) = \mathbf{D}(\mathcal{S}) + \mathbf{E}(\mathcal{S})$, where $\mathbf{D}$ contains the diagonal variance components and $\mathbf{E}$ represents the off-diagonal cross-correlations. 

By defining $\gamma = 1/\sigma^2$ and applying a second-order Taylor expansion to the log-determinant, we obtain the following approximation:
\begin{equation}
\label{eq:logdet_penalty}
\begin{split}
\log \det (\mathbf{I} + \gamma \mathbf{A}(\mathcal{S})) = & \log \det (\mathbf{I} + \gamma \mathbf{D}(\mathcal{S})) \\
& - \frac{1}{2} \text{Tr} \left( [(\mathbf{I} + \gamma \mathbf{D})^{-1} \gamma \mathbf{E}]^2 \right) + \mathcal{O}(\mathbf{E}^3).
\end{split}
\end{equation}
The trace term $\text{Tr}(\cdot)$ inherently acts as an off-diagonal penalty. Since the elements are determined by the spatial kernel $E_{ab} = k(d(s_a, s_b))$, the short-distance bias established in Eq.~\ref{eq:shortpair_bias} implies significantly larger magnitudes in the off-diagonal matrix $\mathbf{E}$ under \textit{TBS}. This strong correlation inevitably leads to a systematic reduction in the expected mutual information:
\begin{equation}
\label{eq:reduction_mutual_information}
\mathbb{E}_{p_C} [I(\mathbf{x}; \mathbf{y} \mid \mathcal{S})] \leq \mathbb{E}_{p_R} [I(\mathbf{x}; \mathbf{y} \mid \mathcal{S})].
\end{equation}

\subsection{Generalization Risk and Intrinsic Difficulty}
According to rate-distortion theory~\cite{cover2012elements}, the reduction in mutual information established in Eq.~\ref{eq:reduction_mutual_information} directly translates to a higher, mathematically unavoidable MSE lower bound. This information loss implies that even an optimal estimator faces a degraded performance floor when confined to trajectory-based observations. 

To explicitly justify the \textit{Shift Degradation} hypothesis, we decompose the excess risk of the RS-trained model $\theta_R$ when evaluated under TBS testing conditions as follows:
\begin{equation}
\label{eq:decomp}
\begin{split}
\mathcal{R}_{p_C}(\theta_R) - \mathcal{R}_{p_R}(\theta_R) = & \underbrace{ \left[ \mathcal{R}_{p_C}(\theta_R) - \mathcal{R}_{p_C}(\theta_C) \right] }_{\text{Distribution Mismatch}} \\
& + \underbrace{ \left[ \mathcal{R}_{p_C}(\theta_C) - \mathcal{R}_{p_R}(\theta_R) \right] }_{\text{Intrinsic Difficulty}}.
\end{split}
\end{equation}
The \textit{Intrinsic Difficulty} term is strictly positive due to the redundancy penalty derived in Eq.~\ref{eq:logdet_penalty}. Simultaneously, the \textit{Distribution Mismatch} term arises from the structural shift in the sampling masks. While these insights are derived from fundamental information theory, their implications naturally extend to deep learning-based RME. When the sampling distribution shifts from uniform i.i.d. patterns to trajectory-constrained clusters, the inductive bias learned by $\theta_R$ becomes completely misaligned with the test-time data manifold. Consequently, the model fails to properly process the highly correlated spatial structures, leading to the severe degradation in generalization observed under \textit{TBS}.

\section{Experiments}\label{sec:exps}
\subsection{Experimental Setup}
\label{sec:exp_setup}

\subsubsection{Dataset Descriptions}
To validate the robustness of RME models across diverse electromagnetic environments, we employ two datasets with distinct propagation characteristics.

\vspace{0.5em} \noindent \textbf{RadioMapSeer~\cite{RadioMapSeer}.} 
This dataset is specifically designed for urban propagation modeling at 5.9 GHz. 
It provides high-fidelity simulations generated via two distinct methodologies: the Dominant Path Model~(DPM) and Intelligent Ray Tracing~(IRT). 
In our experiment, we mainly use the DPM subset.
Each radio map covers a $256 \times 256~\text{m}^2$ area with a spatial resolution of 1~m/pixel. 
The dataset explicitly captures the interaction between signal propagation and urban geometry, with building layouts represented as binary occupancy maps. The dataset contains approximately 56,000 radio maps in total, providing sufficient diversity for training and evaluation.

\vspace{0.5em} \noindent \textbf{SpectrumNet~\cite{SpectrumNet}.} 
As the most extensive open-source 3D radio map repository to date, SpectrumNet spans a vast frequency range from 150 MHz to 22 GHz~(including the 1.5 GHz, 1.7 GHz, and 3.5 GHz bands). 
It is uniquely characterized by its inclusion of diverse geographical terrains, climatic conditions, and material-specific attenuation profiles. 
The full dataset contains approximately 300,000 radio maps. In this work, we focus on the 200 m height layer to investigate the impact of sampling on vertical signal distribution, which corresponds to typical low-altitude aerial or high-rise urban communication scenarios. This subset consists of approximately 100,000 radio maps.
Each radio map has a resolution of $ 128 \times 128$  pixels, with a spatial resolution of 10 m per pixel.

About data partitioning and leakage prevention, we adopt a strict 5:1:4 ratio for training, validation, and testing. 
To ensure the model learns generalizable propagation rules rather than memorizing specific layouts, we implement a layout-aware partition strategy. Notably, in RadioMapSeer, multiple radio maps may share the same building geometry with different transmitter locations. 
We strictly ensure that any building map present in the training or validation sets is entirely excluded from the test set. 
This ``zero-overlap'' constraint on environmental geometries is essential for verifying the model's ability to perform RME across completely unseen urban topographies.

\subsubsection{Evaluation Metrics} 
To quantify the fidelity of the reconstructed radio maps, we employ three complementary metrics that evaluate both point-wise accuracy and structural consistency.
First, the Root Mean Square Error~(RMSE)~$\downarrow$ is used to measure the average pixel-wise deviation between the estimated signal strength and the ground truth, serving as the primary indicator of numerical regression precision. 
Second, we adopt the Structural Similarity Index~(SSIM)~$\uparrow$, which goes beyond simple error residuals by assessing the preservation of luminance, contrast, and spatial structure. 
This is crucial in RME for capturing shadow fading patterns and sharp transitions caused by urban obstacles. 
Finally, the Peak Signal-to-Noise Ratio~(PSNR)~$\uparrow$ is employed to evaluate the dynamic range fidelity of the reconstruction, providing a logarithmic measure of the signal-to-reconstruction-error ratio. 
Collectively, these metrics offer a multi-dimensional perspective on the model's effectiveness in recovering both the global signal distribution and localized propagation details.

\textbf{Comparative Methods and Baselines.} 
Despite significant progress in radio map estimation~(RME), existing research predominantly adopts an idealized Random Sampling~(RS) paradigm, with limited investigation into the impact of sampling distributions on model performance and generalization. To bridge this gap, our comparative analysis focuses on the performance discrepancy between the conventional RS and the more realistic \textit{Trajectory-Based Sampling} (TBS). 
% Within the TBS framework, we define two representative scenarios: first, Mobile Vehicle-TBS~(MVTBS), where the sampling process is dual-constrained by road network topologies and vehicle velocity; second, UAV-TBS, where the trajectory is solely governed by the kinematic velocity constraints of the aerial platform in free space. 
Under the TBS framework, different physical platforms induce distinct forms of trajectory constraints. In particular, we consider two representative instantiations of TBS: (i) vehicle-based TBS, where sampling trajectories are constrained by road network topology and vehicle mobility, and (ii) UAV-based TBS, where trajectories are governed primarily by kinematic velocity constraints in free space.
Furthermore, three road-based trajectory algorithms were introduced from previous papers\cite{vtc}, including a random-based strategy (referred to as RoadRandom) and two genetic algorithm-based strategies ($GA\_c$ and $GA\_s$). The selected trajectories were required to be on the road network and to be continuous. For detailed information about the algorithms, please refer to the papers.

Regarding the RME architectures, we primarily employ RadioUNet, a pioneering framework in this field, as the backbone for our extensive experiments. To verify the architectural universality of the TBS-induced degradation, we extend our evaluation to a wide range of representative models, including classical vision architectures such as U-Net~\cite{unet} and CBAM~\cite{cbam}, as well as state-of-the-art~(SOTA) RME models such as PMNet~\cite{PMNet}, RadioFormer~\cite{radioformer}, and RadioDUN~\cite{radiodun}. In addition to deep learning approaches, we include two traditional spatial-interpolation baselines: Kriging~\cite{Kriging1} and Cubic~\cite{cubic} interpolation methods. This multi-tiered comparative framework is designed to comprehensively reveal the profound impact of sampling-induced distribution shifts across various reconstruction paradigms.

% \subsection{Main Results}
\subsection{Evidence of Sampling Distribution Shift}

% \begin{table}[t]
% \centering
% \caption{Performance comparison under different training-testing sampling distributions on the RadioMapSeer and SpectrumNet datasets.}
% \label{tab:main_results}
% \resizebox{\columnwidth}{!}{%
% \begin{tabular}{ll| ccc c}
% \toprule
% \textbf{Train} & \textbf{Test} & \textbf{RMSE} $\downarrow$ & \textbf{PSNR} $\uparrow$ & \textbf{SSIM} $\uparrow$ & \textbf{$G_{rob}$$\uparrow$} \\ 
% \midrule
% \multicolumn{6}{l}{\cellcolor[HTML]{F2F2F2}\textbf{Dataset: RadioMapSeer}} \\ \addlinespace[0.1cm]
%  Random & Random & \textbf{0.0305} & \textbf{30.39} & \textbf{0.9506} & / \\   % $Random \rightarrow  Random$
% \rowcolor[HTML]{D9D9D9} Random & ST-TBS & 0.0775 & 22.30 & 0.9174 & / \\
% \cmidrule(lr){1-6}
% ST-TBS & ST-TBS & 0.0570 & 24.88 & 0.9349 & 26.45\% \\
% ST-TBS & Random & 0.0310 & 30.43 & 0.9478 & 60.00\% \\ 
% % \addlinespace[0.2cm]
% \midrule
% \multicolumn{6}{l}{\cellcolor[HTML]{F2F2F2}\textbf{Dataset: SpectrumNet}} \\ \addlinespace[0.1cm]
% Random & Random & \textbf{0.0391} & \textbf{28.24} & \textbf{0.8408} & / \\
% \rowcolor[HTML]{D9D9D9} Random & ST-TBS & 0.2632 & 11.60 & 0.5332 & / \\
% \cmidrule(lr){1-6}
% ST-TBS & \underline{ST-TBS} & \underline{0.0571} & \underline{24.97} & \underline{0.8142} & \underline{78.31\%} \\ 
% ST-TBS & Random & 0.0545 & 25.42 & 0.8196 & 79.29\% \\
% \bottomrule
% \end{tabular}%
% }
% \end{table}

\begin{table}[t]
\centering
\caption{Performance comparison under different training-testing sampling distributions on the RadioMapSeer and SpectrumNet datasets.}
\label{tab:main_results}
\resizebox{\columnwidth}{!}{%
\begin{tabular}{l| ccc c}
\toprule
\textbf{Train $\rightarrow$ Test} & \textbf{RMSE} $\downarrow$ & \textbf{PSNR} $\uparrow$ & \textbf{SSIM} $\uparrow$ & \textbf{$G_{rob}$$\uparrow$} \\ 
\midrule
\multicolumn{5}{l}{\cellcolor[HTML]{F2F2F2}\textbf{Dataset: RadioMapSeer}} \\ \addlinespace[0.1cm]
% $\text{Random} \rightarrow \text{Random}$
Random $\rightarrow$ Random & \textbf{0.0305} & \textbf{30.39} & \textbf{0.9506} & / \\  
\rowcolor[HTML]{D9D9D9}
Random $\rightarrow$ ST\text{-}TBS & 0.0775 & 22.30 & 0.9174 & / \\
\cmidrule(lr){1-5}
ST\text{-}TBS $\rightarrow$ ST\text{-}TBS & 0.0570 & 24.88 & 0.9349 & 26.45\% \\
ST\text{-}TBS $\rightarrow$ Random & 0.0310 & 30.43 & 0.9478 & 60.00\% \\ 
\midrule
\multicolumn{5}{l}{\cellcolor[HTML]{F2F2F2}\textbf{Dataset: SpectrumNet}} \\ \addlinespace[0.1cm]
Random $\rightarrow$ Random & \textbf{0.0391} & \textbf{28.24} & \textbf{0.8408} & / \\
\rowcolor[HTML]{D9D9D9}
Random $\rightarrow$ ST\text{-}TBS & 0.2632 & 11.60 & 0.5332 & / \\
\cmidrule(lr){1-5}
\underline{ST\text{-}TBS $\rightarrow$ ST\text{-}TBS} & \underline{0.0571} & \underline{24.97} & \underline{0.8142} & \underline{78.31\%} \\ 
ST\text{-}TBS $\rightarrow$ Random & 0.0545 & 25.42 & 0.8196 & 79.29\% \\
\bottomrule
\end{tabular}%
}
\end{table}

\begin{table}[t]
\centering
\caption{
Performance comparison under different trajectory-based sampling strategies on the RadioMapSeer dataset.} 
\label{tab:main_results_expand}
\resizebox{\columnwidth}{!}{%
\begin{tabular}{ll| ccc c}
\toprule
\textbf{Train} & \textbf{Test} & \textbf{RMSE} $\downarrow$ & \textbf{PSNR} $\uparrow$ & \textbf{SSIM} $\uparrow$ & \textbf{$G_{rob}$$\uparrow$} \\ 
\midrule
\multicolumn{6}{l}{\cellcolor[HTML]{F2F2F2}\textbf{Dataset: RadioMapSeer}} \\ \addlinespace[0.1cm]
\rowcolor[HTML]{D9D9D9} Random & ST-TBS & 0.0775 & 22.30 & 0.9174 & / \\
Random & RoadRandom & 0.0760 & 23.36 & 0.8983 & 1.94\% \\
Random & GA\_c & 0.0491 & 26.49 & 0.9261 & 36.65\% \\
Random & GA\_s & 0.0419 & 27.65 & 0.9322 & 45.94\% \\
\cmidrule(lr){1-6}
ST-TBS & RoadRandom & 0.0526 & 26.28 & 0.9237 & 32.13\% \\
ST-TBS & GA\_c & 0.0387 & 28.39 & 0.9352 & 50.06\% \\
ST-TBS & \underline{GA\_s} & \underline{0.0379} & \underline{28.56} & \underline{0.9358} & \underline{51.10\%} \\ 
\bottomrule
\end{tabular}%
}
\end{table}

\subsubsection{Main Experimental Results and Performance Analysis}

To evaluate the impact of sampling paradigms on radio field reconstruction and the efficacy of our proposed training strategy, we conduct comprehensive experiments using the \textbf{RadioUNet} architecture as the backbone. The number of sampling points is fixed at $N=50$ for all scenarios to ensure a fair comparison of information efficiency. The experimental results are summarized in Table~\ref{tab:main_results} and Table~\ref{tab:main_results_expand}, where Table~\ref{tab:main_results} highlights the impact of sampling distribution mismatch under different training–testing configurations, and Table~\ref{tab:main_results_expand} provides additional analysis on different trajectory-based sampling strategies, including random trajectories and optimized paths (e.g., GA-c and GA-s). 
\textbf{Bold} indicates the best empirical performance under unconstrained random sampling, while \underline{underlined} values denote the best performance under physical trajectory constraints. $G_{rob}$ represents the relative RMSE reduction compared to the most degraded (Random$\rightarrow$ST-TBS) baseline under the same dataset. Cells with a \colorbox[HTML]{D9D9D9}{gray background} highlight the performance collapse under sampling distribution shifts.

\vspace{0.5em} \noindent \textbf{Robustness Against Sampling Distribution Shift.}  

As shown in Table~\ref{tab:main_results}, a significant performance gap is observed when the training and testing sampling distributions are inconsistent. As highlighted by the \colorbox[HTML]{D9D9D9}{gray rows}, the \textit{Random-trained} baseline experiences a catastrophic performance collapse when evaluated under Trajectory-Based Sampling (TBS). Specifically, on the SpectrumNet dataset, the RMSE surges from 0.0391 to 0.2632, a near seven-fold increase in error. This result highlights that trajectory-constrained sampling introduces strong spatial correlation and local clustering, violating the $i.i.d.$ assumption of random sampling and leading to a critical distribution shift. Similar trends are consistently observed on the RadioMapSeer dataset, confirming that this issue is not dataset-specific.

\begin{figure}[t]
\centering
\includegraphics[width=\columnwidth]{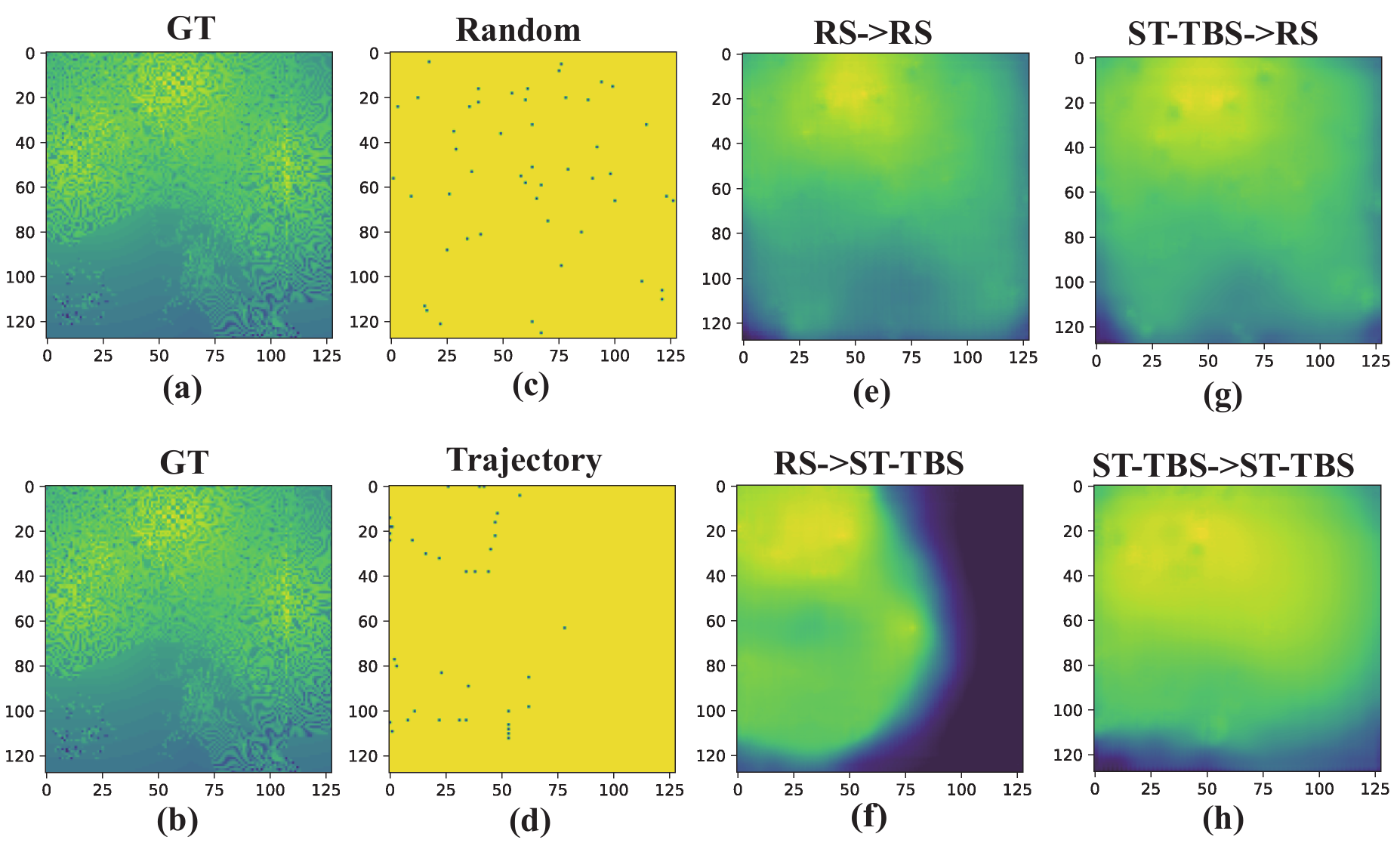}
\caption{
Qualitative reconstruction comparison on the SpectrumNet dataset using RadioUNet under different training-testing sampling configurations.
The first column shows the ground truth maps. The second column visualizes the corresponding sampling patterns, including random sampling (RS) and stochastic-triggered trajectory-based sampling (ST-TBS). The remaining columns present reconstruction results under four training-testing configurations: RS$\to$RS, ST-TBS$\to$RS, RS$\to$ST-TBS, and ST-TBS$\to$ST-TBS. The content before the arrow represents the trained model, and the content after the arrow represents the sampling method.}
%When the training and testing sampling distributions are matched (RS$\to$RS and ST-TBS$\to$ST-TBS), accurate reconstructions are achieved. In contrast, under mismatched conditions, the RS-trained model fails to recover large unsampled regions when evaluated on trajectory-based inputs (RS$\to$ST-TBS), whereas the ST-TBS-trained model exhibits improved spatial generalization and produces more coherent reconstructions.
\label{fig:spectrumnet exp}
\end{figure}

To further investigate the severe performance degradation observed on the SpectrumNet dataset, we provide qualitative visualization results in Fig.~\ref{fig:spectrumnet exp}.
It is important to note that although the UAV follows a continuous trajectory, measurements are collected at discrete intervals, resulting in sparsely distributed sampling points along the path.
A key observation is that the model trained with random sampling (RS-trained) exhibits a strong dependence on local sample coverage. 
As shown in Fig.~\ref{fig:spectrumnet exp}(f), when evaluated under trajectory-based sampling (ST-TBS), large regions without direct observations cannot be effectively reconstructed. 
This is because the RS-trained model implicitly learns a point-wise interpolation behavior, 
where information propagation is primarily driven by nearby sampled points. 
Consequently, regions lacking sufficient sampling support remain poorly estimated.

In contrast, the model trained with trajectory-based sampling (TBS-trained) demonstrates significantly improved robustness. As illustrated in Fig.~\ref{fig:spectrumnet exp}(h), even in sparsely observed regions, the model is able to infer plausible signal distributions. This suggests that TBS-aware training encourages the model to exploit spatial correlations and learn more global structural patterns, rather than relying solely on local interpolation from independently distributed samples. 

This difference is particularly pronounced in SpectrumNet, where the propagation environment exhibits smooth and weakly structured spatial variations. In such scenarios, effective reconstruction relies heavily on global information inference. Therefore, the clustered sampling pattern in TBS, when properly incorporated during training, can guide the model to better generalize under realistic sensing conditions.

\begin{figure*}[t]
\centering
\includegraphics[width=0.95\textwidth]{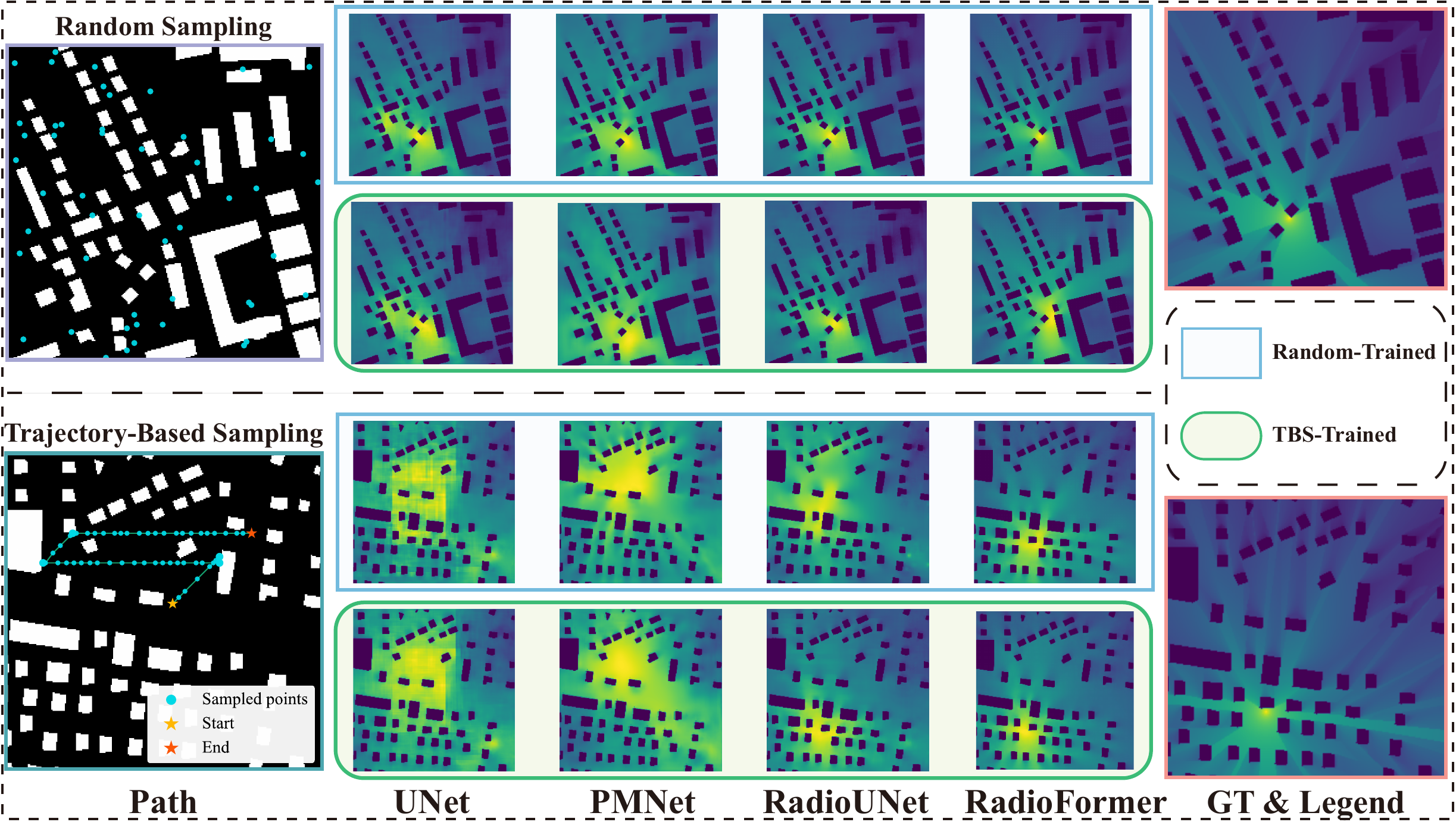}
\caption{Qualitative comparison of reconstruction performance under different sampling strategies and training paradigms on the RadioMapSeer dataset. The left column illustrates the sampling patterns, including Random Sampling (top) and Trajectory-Based Sampling (bottom). For each sampling strategy, reconstruction results from multiple models (UNet, PMNet, RadioUNet, and RadioFormer) are presented under two training schemes: the blue rectangular box represents the Random-trained model, while the green rounded box represents the TBS-trained model. The rightmost column shows the corresponding ground truth.}
\label{fig:multi_model_comparison}

\end{figure*}

\vspace{0.5em} \noindent \textbf{Efficacy of Training-Aware Strategy.}  

Table~\ref{tab:main_results} further shows that the proposed ST-TBS training strategy effectively mitigates this degradation. On SpectrumNet, the ST-TBS-trained model achieves a robustness gain ($G_{rob}$) of 79.29\%, substantially recovering performance under trajectory-based testing.
Moreover, on RadioMapSeer, the ST-TBS-trained model maintains strong performance across both trajectory-based and random testing. Notably, it achieves an RMSE of 0.0310 under Random testing, which is close to the domain-optimal Random→Random result (0.0305). This indicates that trajectory-aware training improves generalization without sacrificing performance under ideal sampling conditions.

\vspace{0.5em} \noindent \textbf{Impact of Trajectory Patterns and Path Strategies.}  

To further analyze the effect of different trajectory patterns, Table~\ref{tab:main_results_expand} compares multiple trajectory-based sampling strategies. Due to the absence of street network topology in the SpectrumNet dataset, path-dependent baselines are omitted. The results show that trajectory design plays an important role in reconstruction performance. In particular, optimized trajectories such as GA-c and GA-s provide better spatial coverage and lead to improved performance compared to simple path-consistent sampling.

However, regardless of the specific trajectory strategy, models trained with ST-TBS consistently outperform the ones trained with random sampling. This demonstrates that aligning the training process with trajectory-constrained sampling is more critical than optimizing individual path instances.

\vspace{0.5em} \noindent \textbf{Insights on Sampling and Deployment.}   

A key finding is that our previous work, \textbf{GA\_s} (Genetic Algorithm-based Sampling), consistently achieves superior performance across all evaluation metrics, yielding the lowest root mean square error (RMSE) of \textbf{0.0379}. This result reveals an important insight: although robust training strategies such as ST-TBS serve as essential safeguards, the deliberate optimization of the sampling pattern itself (i.e., GA-s) exerts a stronger influence on overall model performance in practical engineering applications. Consequently, advancing sampling paradigms warrants equal emphasis alongside model architecture design in pursuit of high-fidelity radio environmental mapping. 

By comparing Random$\to$TBS and TBS$\to$TBS configurations, we can disentangle the impact of distribution mismatch and information insufficiency. The performance gap between Random$\to$TBS and TBS$\to$TBS reflects the additional degradation caused by distribution shift, while the gap between TBS$\to$TBS and Random$\to$Random captures the intrinsic difficulty due to reduced information diversity. This result provides practical guidance for UAV deployment, suggesting that trajectory-based sensing systems may require a larger sampling budget to compensate for information redundancy.

\begin{table}[t]
\centering
\caption{RMSE comparison across models under sampling distribution mismatch scenarios on the RadioMapSeer dataset.}
\label{tab:multi_models_comparison}
\renewcommand{\arraystretch}{1.1} 
\begin{tabular}{l c c c c}
\toprule
Model & R $\rightarrow$ R & R $\rightarrow$ T & T $\rightarrow$ R & T $\rightarrow$ T \\
\midrule
Kriging            & 0.1791 & 0.2319 & 0.1791 & 0.2319 \\
Cubic              & 0.1816 & 0.2288 & 0.1816 & 0.2288 \\
\midrule
\textit{PMNet}~\cite{PMNet}     & 0.0370 & 0.0853 & 0.0753 & 0.0974 \\
\textit{UNet}~\cite{unet}      & 0.0357 & 0.0855 & 0.0398 & 0.0835 \\
\textit{CBAM}~\cite{cbam}      & 0.0336 & 0.0913 & 0.0394 & 0.0737 \\
\textit{RadioUNet}~\cite{RadioUNet} & 0.0305 & \bfseries 0.0775 &  0.0310 & 0.0570 \\
\textit{RadioDUN}~\cite{radiodun}  & 0.0285 & 0.0825 & \bfseries 0.0301 & 0.0632 \\
\textit{RadioFormer}~\cite{radioformer}& \bfseries 0.0186 & 0.0989 & 0.0349 & \bfseries 0.0552 \\
\bottomrule
\end{tabular}
\end{table}

\subsubsection{Cross-Architecture Robustness to Sampling Distribution Shift} 

To further investigate whether the performance degradation under sampling distribution shifts is model-dependent, we evaluate several representative reconstruction networks on the RadioMapSeer dataset, including UNet, CBAM, PMNet, RadioFormer, and RadioUNet. The quantitative reconstruction results under different training and testing sampling configurations are summarized in Table~\ref{tab:multi_models_comparison}. 
We consider four train--test configurations, including 
(i) Random$\rightarrow$Random~(R$\rightarrow$R), 
(ii) Random$\rightarrow$ST-TBS~(R$\rightarrow$T), 
(iii) ST-TBS$\rightarrow$Random~(T$\rightarrow$R), and 
(iv) ST-TBS$\rightarrow$ST-TBS~(T$\rightarrow$T).

\vspace{0.5em} \noindent \textbf{Performance under matched sampling distributions.}
When the training and testing sampling distributions are consistent (R$\to$R or T$\to$T), all models achieve relatively low reconstruction errors, as shown in Table~\ref{tab:multi_models_comparison}. 
For example, RadioFormer achieves an RMSE of 0.0186 under R$\to$R, while RadioUNet also obtains competitive performance. 
Similarly, under T$\to$T, most models maintain stable reconstruction accuracy.
These results indicate that modern deep reconstruction models are capable of effectively learning the mapping between sparse measurements and full radio maps when the sampling distribution during training matches that of deployment.

\vspace{0.5em} \noindent \textbf{Severe degradation under sampling distribution shifts.}
However, when the training and testing sampling distributions are mismatched, the reconstruction performance deteriorates significantly across all architectures. 
For instance, when models trained with random sampling are evaluated under trajectory-based sampling (R$\to$T), the RMSE increases substantially. 
RadioFormer, for example, degrades from 0.0186 to 0.0989. 
A similar degradation is observed in the opposite setting (T$\to$R), indicating that sampling distribution mismatch leads to severe generalization issues for existing reconstruction networks.

\vspace{0.5em} \noindent \textbf{Architecture improvement does not resolve the mismatch.}
Interestingly, this degradation consistently appears across different model architectures, including CNN-based models (UNet, CBAM), hybrid architectures (RadioUNet), and transformer-based models (RadioFormer). 
Despite their differences in architectural design and representation capacity, none of these models demonstrates strong robustness to sampling distribution shifts. This suggests that the performance degradation is not primarily caused by insufficient model capacity, but rather by the inherent mismatch between the sampling strategy used during training and the trajectory-constrained sampling encountered during deployment.

\vspace{0.5em} \noindent \textbf{Visualization results.}
Fig.~\ref{fig:multi_model_comparison} presents qualitative reconstruction results under different train-test sampling configurations. When the sampling distributions are consistent (R$\to$R or T$\to$T), the reconstructed radio maps closely match the ground truth. In contrast, under distribution mismatch scenarios (R$\to$T or T$\to$R), noticeable reconstruction artifacts and spatial distortions appear across all models.

\begin{figure}[t]
\centering
\includegraphics[width=\columnwidth]{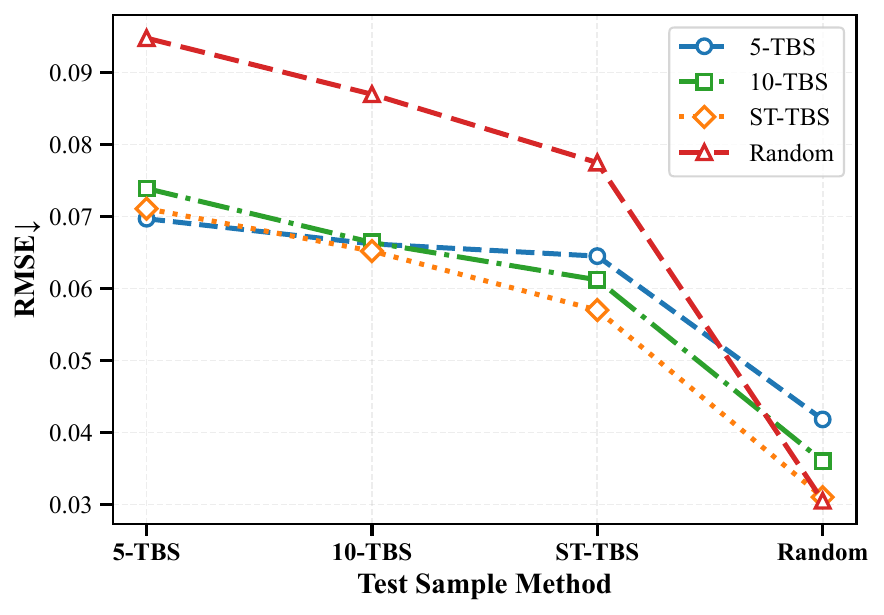}
\caption{Sensitivity and robustness analysis across various training and testing sampling paradigms. The horizontal axis represents different sampling methods employed during testing, including three Trajectory-Based Sampling (TBS) variants ($\ell=5$, $\ell=10$, and ST-TBS) and a Random Sampling baseline. The curves illustrate the RMSE performance of models optimized under different training regimes, with the vertical displacement between curves quantifying robustness against sampling-induced distribution shifts.} 
\label{fig:ablation_line}
\end{figure}

\begin{figure*}
\centering
\includegraphics[width=0.95\textwidth]{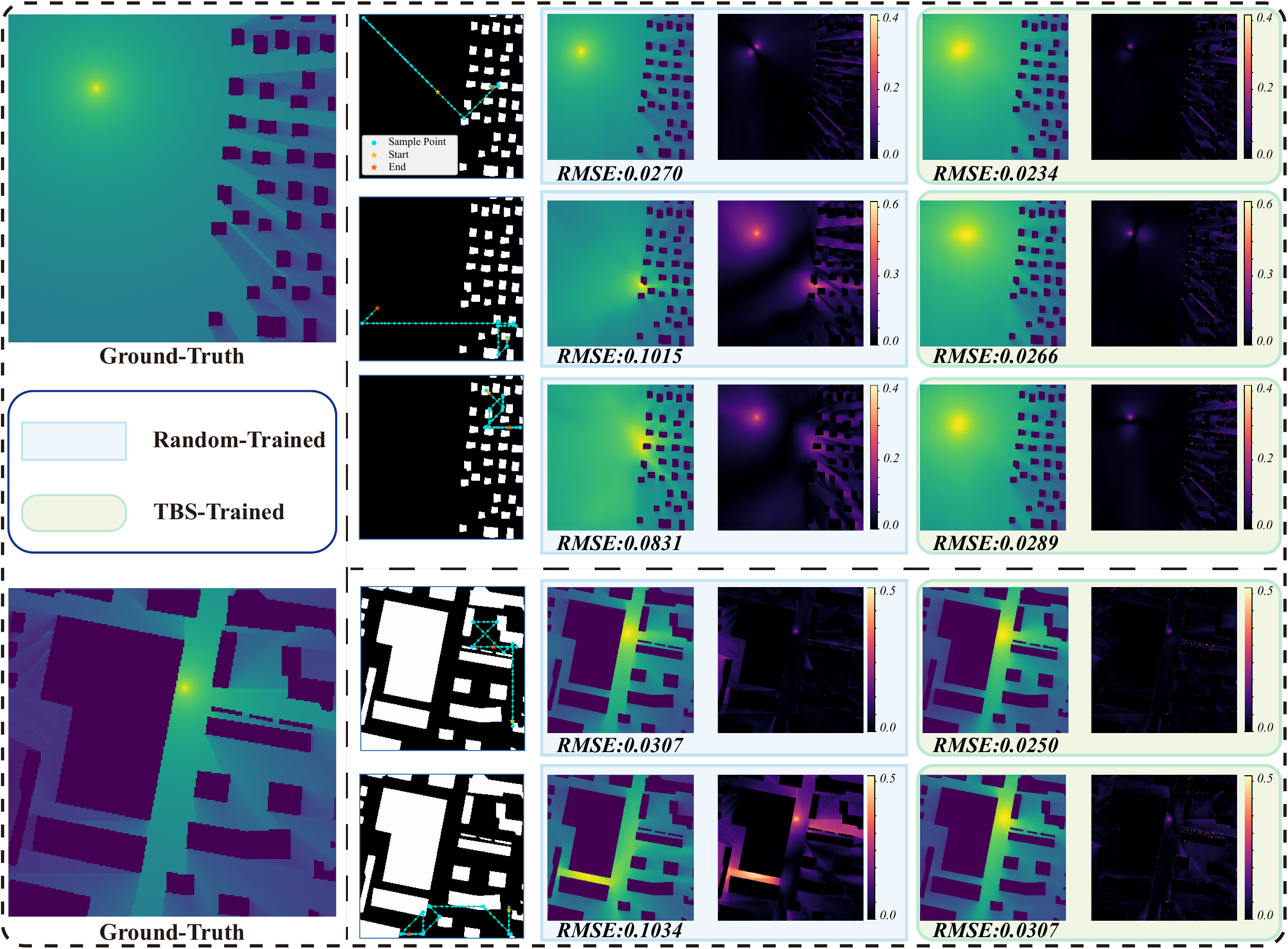}
\caption{
Ablation study on the impact of trajectory patterns on reconstruction performance across different environments. Two representative radio maps are considered (top and bottom), with corresponding ground-truth distributions shown on the left. 
For each map, multiple trajectory patterns are illustrated, including three distinct paths in the first scenario and two paths in the second scenario.
For each trajectory, reconstruction results are presented using models trained under random sampling (blue rectangle) and trajectory-based sampling (green rounded rectangle), along with the corresponding error maps and RMSE values.}

\label{fig:multi_paths_comparison}
\end{figure*}

\subsection{Ablation Study}

\subsubsection{Multi-Dimensional Analysis of Sampling Dynamics}

To evaluate the proposed trajectory-based sampling (TBS) framework and the effectiveness of the training strategies, we conduct an ablation study using RadioUNet as the baseline model on the RadioMapSeer dataset.

Fig.~\ref{fig:ablation_line} summarizes the experimental results, providing quantitative evidence of how trajectory-induced kinematic constraints affect reconstruction performance. Specifically, we analyze the results from two complementary perspectives to better understand the impact of sampling dynamics on estimation accuracy.

\vspace{0.5em} \noindent \textbf{Test-Time Sensitivity.}
Regardless of the training regime, all models exhibit a monotonic performance degradation as the testing sampling paradigm transitions from independent points (\textit{Random}) to highly constrained trajectories ($\ell=5$). 
As illustrated by the curves, the RMSE of the random-trained baseline increases significantly from 0.0305 to 0.0948 when shifting from the RS to the $\ell=5$ TBS testing scenario. 
This consistent upward trend across all models confirms that tighter motion constraints (smaller $\ell$) exacerbate spatial redundancy and reduce the effective information gain per sample, thereby increasing the intrinsic difficulty of global field estimation.

\vspace{0.5em} \noindent \textbf{Training Strategy Robustness.} 
Comparing the vertical displacement of the curves reveals the efficacy of the proposed TBS-aware training in mitigating distribution shifts. 
The \textit{Random-trained} model~(represented by the top-most curve) consistently exhibits the highest RMSE across all TBS test scenarios, demonstrating its failure to generalize to path-dependent observations. 
In contrast, the model trained with the \textit{Stochastic Triggering}~(ST-TBS) strategy~(orange line) demonstrates superior robustness. 
Notably, the ST-TBS model achieves the lowest RMSE under all TBS-constrained testing scenarios, while maintaining a comparable RMSE of 0.0310 under random sampling, which is close to the domain-optimal performance of the Random-trained model~(0.0305). 
The small performance gap suggests that introducing stochastic perturbations during training mitigates overfitting to specific sampling patterns and promotes a more generalizable inductive bias, enabling robust performance across both structured trajectory-based and independent random sampling regimes.
This result highlights the importance of matching training sampling distributions with real-world acquisition constraints.

\subsubsection{Generalization Across Trajectories}
To further evaluate the robustness of the proposed approach under varying trajectory patterns, we conduct this study using the RadioFormer model on the RadioMapSeer dataset, as shown in Fig.~\ref{fig:multi_paths_comparison}.
Two representative radio maps are considered (top and bottom), each associated with multiple trajectory patterns exhibiting different spatial coverage and sampling characteristics. Specifically, three distinct trajectories are evaluated on the first map, while two additional trajectories are considered on the second map, covering scenarios ranging from relatively uniform coverage to highly clustered observations.

\vspace{0.5em} \noindent \textbf{Sensitivity to Trajectory Patterns.} 
From Fig.~\ref{fig:multi_paths_comparison}, a clear contrast can be observed between models trained under random sampling and those trained with trajectory-based sampling. The Random-trained model is highly sensitive to the specific trajectory pattern. When the trajectory exhibits limited spatial coverage or strong local clustering, it fails to capture the global signal structure, leading to large reconstruction errors (e.g., RMSE values of 0.1015 and 0.1034).
In contrast, the TBS-trained model demonstrates stable performance across all trajectory patterns. Even under challenging sampling configurations, it consistently maintains low estimation errors (typically around 0.026-0.030), indicating strong generalization under trajectory-constrained observations.

\vspace{0.5em} \noindent \textbf{Impact of Sampling Structure on Model Behavior.} 
This behavior highlights a fundamental difference in how the two models utilize observations. The Random-trained model primarily relies on local interpolation from independently distributed samples, making it vulnerable to structured or uneven sampling. In comparison, the TBS-trained model learns to exploit spatial correlations along trajectories and perform more global inference, enabling robust estimation even under constrained sampling conditions. Overall, these results confirm that the proposed TBS-aware training strategy significantly improves cross-trajectory generalization and avoids overfitting to specific sampling patterns.

\subsubsection{Continuous Transition Between Random and Trajectory-Based Sampling}
To further investigate how reconstruction performance varies with sampling structure, we evaluate hybrid sampling distributions by varying the path ratio $\alpha$, which denotes the proportion of trajectory-based samples. For each test instance, the total number of samples is fixed, while $\alpha$ is varied to interpolate between independent random sampling ($\alpha=0$) and fully trajectory-based sampling ($\alpha=1$).

\begin{figure}[t]
\centering
\includegraphics[width=\columnwidth]{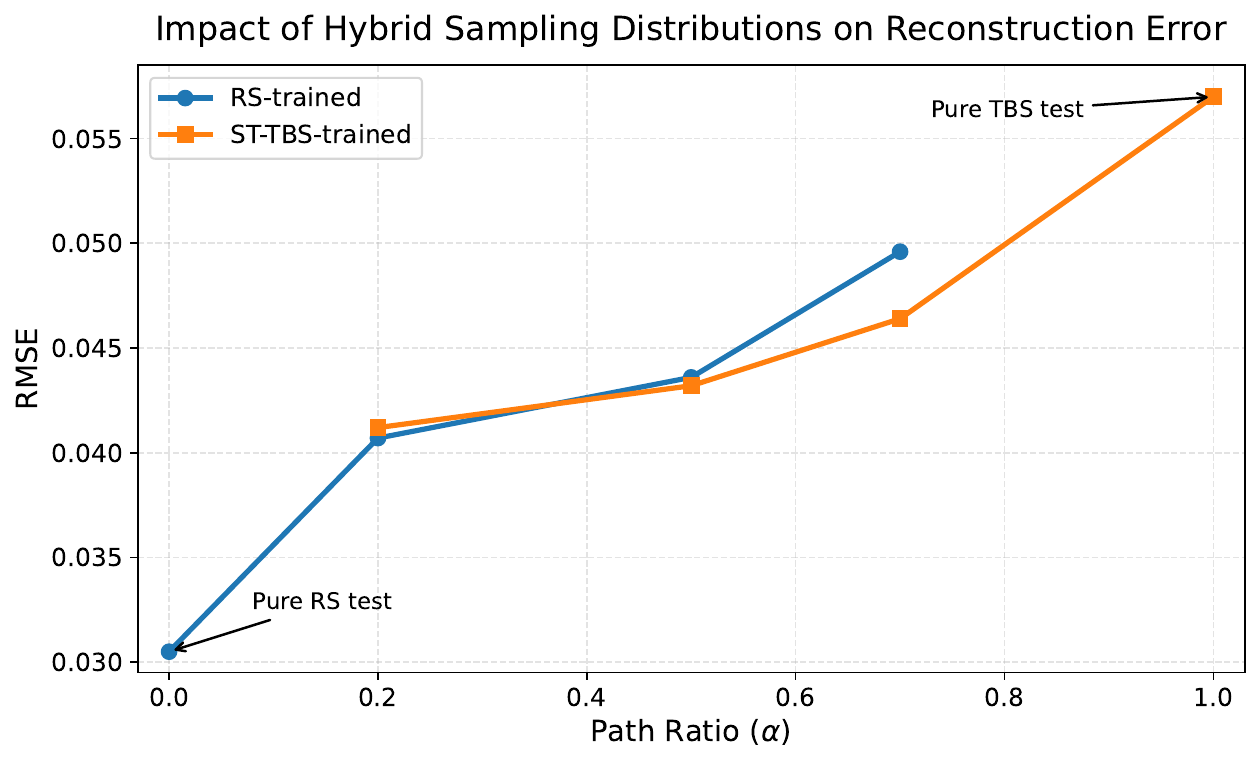}
\caption{Reconstruction performance under hybrid sampling distributions with varying path ratio $\alpha$.
The parameter $\alpha$ denotes the proportion of trajectory-based samples, where $\alpha=0$ corresponds to random sampling and $\alpha=1$ corresponds to trajectory-based sampling.} 
\label{fig:hybird_sample}
\end{figure}

\vspace{0.5em} \noindent \textbf{Monotonic Performance Degradation.} 
As shown in Fig.~\ref{fig:hybird_sample}, the reconstruction error increases monotonically with $\alpha$, indicating that performance degradation evolves continuously with the degree of trajectory-induced spatial correlation. The specific values can be found in Table~\ref{tab:hybrid_sample}. This observation confirms that the impact of sampling distribution is not binary, but instead varies progressively as the sampling pattern transitions from independent to trajectory-constrained observations.

\vspace{0.5em} \noindent \textbf{Training-Testing Distribution Alignment.} 
Furthermore, models trained with ST-TBS exhibit improved robustness under higher $\alpha$ values, outperforming randomly trained models when trajectory-based components dominate. In contrast, randomly trained models achieve slightly better performance when $\alpha$ is small. These results highlight that reconstruction performance is primarily governed by the alignment between training and testing sampling distributions, rather than model capacity alone.
This experiment provides a continuous perspective on sampling-induced distribution shifts, bridging the gap between idealized random sampling and fully constrained trajectory-based sensing scenarios.

\begin{table}[t]
\centering
\caption{
Impact of hybrid sampling distributions on reconstruction performance. 
The path ratio $\alpha$ denotes the proportion of trajectory-based samples. 
}
\label{tab:hybrid_sample}
\begin{tabular}{c|c|c}
\toprule
\textbf{Path Ratio ($\alpha$)} & \textbf{Random-trained} & \textbf{ST-TBS-trained} \\
\midrule
0.0 & 0.0305 & - \\
0.2 & 0.0407 & 0.0412 \\
0.5 & 0.0436 & 0.0432 \\
0.7 & 0.0496 & 0.0464 \\
1.0 & - & 0.0570 \\
\bottomrule
\end{tabular}
\end{table}

\section{Conclusion}
\label{sec:conclusion}
In this paper, we identified and analyzed the simulation-to-deployment gap in learning-based radio map estimation caused by sampling distribution mismatch between random and trajectory-based sampling. Our results show that trajectory-based sampling introduces strong spatial correlations and redundancy, which significantly degrade reconstruction performance.
Extensive experiments on the RadioMapSeer and SpectrumNet datasets show that models trained under random sampling suffer severe performance degradation under trajectory-based observations. Moreover, this degradation evolves continuously with the degree of trajectory-induced correlation, highlighting that aligning training and testing sampling distributions is critical for robust reconstruction. 
Notably, stronger models are found to be more sensitive to sampling distribution mismatch, indicating that increasing model capacity alone cannot resolve this issue. 
Overall, these findings suggest that the mismatch between training and deployment sampling processes  constitutes a fundamental limitation in current learning-based RME frameworks.
Future work will explore adaptive sensing strategies that jointly optimize UAV trajectory planning and reconstruction performance.

% \section*{Acknowledgments}
% This should be a simple paragraph before the References to thank those individuals and institutions who have supported your work on this article.

% cite
% \newpage
\bibliographystyle{IEEEtran}
\bibliography{IEEEabrv,cite}

\end{document}